\documentclass{llncs}
\usepackage{amssymb}
\usepackage{amsmath}

\usepackage{graphicx}
\usepackage[utf8]{inputenc}
\usepackage{color}

\usepackage{graphicx}
\usepackage{tabularx}
\usepackage{xcolor}
\usepackage{float}
\usepackage{rotating}
\usepackage{xstring} 
\usepackage{algorithm}
\usepackage[noend]{algorithmic}

\usepackage[colorlinks,citecolor=blue]{hyperref}

\newcommand{\isdraft}{\boolean{false}}  
\usepackage{ifthen}
\providecommand{\isdraft}{\boolean{false}}
\ifthenelse{\isdraft}{}{\newcommand{\markupdraft}[2]{}}
\providecommand{\markupdraft}[2]{
    \ifthenelse{\equal{#1}{display}}{#2}{}
    \ifthenelse{\equal{#1}{color}}{\color{#2}}{}
}
\newcommand{\notecolored}[3][]{\markupdraft{display}{{\color{#2}\noindent[Note (#1): #3]}}}
\newcommand{\newcolored}[3][]{{\markupdraft{color}{#2}#3}
    \ifthenelse{\equal{#1}{}}{}{\markupdraft{display}{{\color{yellow!70!black}[#1]}}}} 

\providecommand{\del}[2][]{{\markupdraft{display}{{\color{yellow!80!black}[removed: "#2"[#1]]}}}} 

\newcommand{\niko}[1]{\notecolored[Niko]{green!90!yellow!80!black}{#1}} 
\newcommand{\ilya}[1]{\notecolored[Ilya]{red!50!yellow!50!black}{#1}} 

\newcommand{\R}{\mathbb{R}}

\newcommand{\C}{\textbf{\textit{C}}}
\newcommand{\m}{\textbf{\textit{m}}}

\newcommand{\x}{\textbf{\textit{x}}}

\newcommand{\dd}{n}
\newcommand{\vc}[1]{\textit{\textbf{#1}}}
\def\ONE{{\rm 1\hspace{-0.50ex}1}}

\def\NormOI{{\mathcal N}  \hspace{-0.13em}\left({\ma{0}, \ensuremath{\ma{I}}\,}\right)}
\def\ONE{{\rm 1\hspace{-0.80ex}1}}
\def\Id{\ensuremath{\ma{I}}}

\def\x{\vc{x}}

\def\m{\vc{m}}

\newcommand{\ma}[1]{\mathchoice{\mbox{\boldmath$\displaystyle#1$}}
  {\mbox{\boldmath$\textstyle#1$}} {\mbox{\boldmath$\scriptstyle#1$}}
  {\mbox{\boldmath$\scriptscriptstyle#1$}}}
\renewcommand{\ma}[1]{\mathnormal{\mathbf{#1}}}
\newcommand{\mstr}[1]{\mathrm{#1}}

\definecolor{mygray}{rgb}{.8,.8,.8}

\begin{document}
\title{
Maximum Likelihood-based Online Adaptation of Hyper-parameters in CMA-ES
}

\author{
Ilya Loshchilov\inst{1} 
\and Marc Schoenauer\inst{2}$^{,3}$ \and Mich{\`e}le Sebag\inst{3}$^{,2}$
\and Nikolaus~Hansen\inst{2}$^{,3}$}

\institute{
Laboratory of Intelligent Systems, \\ \'Ecole Polytechnique F\'ed\'erale de Lausanne, Switzerland,
  {\tt Ilya.Loshchilov@epfl.ch}
\and
TAO Project-team, INRIA Saclay - \^Ile-de-France
\and Laboratoire de Recherche en Informatique 
(UMR CNRS 8623)\\
Universit\'e Paris-Sud, 91128 Orsay Cedex, France\\
{\tt FirstName.LastName@inria.fr}
}

\maketitle

\newcommand{\Free}{self}

\begin{abstract}

The Covariance Matrix Adaptation Evolution Strategy (CMA-ES) is widely accepted as a robust  derivative-free continuous optimization algorithm for non-linear and non-convex optimization problems. 
CMA-ES is well known to be almost parameterless, meaning that only one hyper-parameter, the population size, is proposed to be tuned by the user. In this paper, we propose a principled approach called \Free-CMA-ES to achieve the online adaptation of CMA-ES hyper-parameters in order to improve its overall performance. Experimental results show that for larger-than-default population size, the default settings of hyper-parameters of CMA-ES are far from being optimal, and that \Free-CMA-ES allows for dynamically approaching optimal settings. 
\end{abstract}

\section{Introduction}

The Covariance Matrix Adaptation Evolution Strategy (CMA-ES \cite{2003HansenCMA}) 
is a continuous optimizer 
which only exploits the ranking of estimated candidate solutions to approach the optimum of
 an objective function $f: \R^n \rightarrow \R$. 
CMA-ES is also invariant w.r.t. affine transformations of the decision space, explaining the known robustness of the algorithm. 
An important practical advantage of CMA-ES is that all hyper-parameters thereof are defined by default with respect to the problem dimension $n$. Practically, only the population size $\lambda$ is suggested to be tuned by the user, e.g. when a parallelization of the algorithm is considered or  the problem at hand is known to be multi-modal and/or noisy \cite{2005AugerIPOP,2010HansenIPOPaCMAonNoisy}. Other hyper-parameters have been provided robust default settings (depending on $n$ and $\lambda$), in the sense that their offline tuning allegedly hardly improves the CMA-ES performance for unimodal functions. In the meanwhile, for multi-modal functions it is suggested that the overall performance can be significantly improved by offline tuning of $\lambda$  and multiple stopping criteria \cite{2010SmitEibenCMAEStuning,2013LiaoStutzleCEC}. Additionally, it is shown that CMA-ES can be 
improved by a factor up to 5-10 by the use of surrogate models on unimodal ill-conditioned functions \cite{2013LoshchilovGECCO}. This suggests that the CMA-ES performance can be improved by better exploiting the information in the evaluated samples $(\vc{x},f(\vc{x}))$.  

This paper focuses on the automatic online adjustment of the CMA-ES hyper-parameters. The proposed approach, called \Free-CMA-ES, relies on a second CMA-ES instance operating on the hyper-parameter space of the first CMA-ES, and aimed at increasing the likelihood of generating the most successful samples $\vc{x}$ in the current generation. 
The paper is organized as follows. Section \ref{section:CMA} describes the original $(\mu / \mu_w,\lambda)$-CMA-ES.
\Free-CMA-ES is described in Section \ref{section:Approach} and its experimental validation is discussed in Section \ref{sectio:experiments} comparatively to related work. Section \ref{section:conclusion} concludes the paper.

\section{Covariance Matrix Adaptation Evolution Strategy}
\label{section:CMA}
The Covariance Matrix Adaptation Evolution Strategy \cite{1996HansenCMAES,2001HansenCMA,2003HansenCMA} is acknowledgedly the most popular and the most efficient Evolution Strategy algorithm. 

The original ($\mu/\mu_{w},\lambda$)-CMA-ES (Algorithm \ref{CMAdefault}) proceeds as follows.
At the $t$-th iteration, a Gaussian distribution ${\mathcal N}  \hspace{-0.13em}\left({\vc{m}^t,{\sigma^t}^2 {\C}^t}\right)$ is used to generate $\lambda$ candidate solution $\vc{x}_k \in \mathbb{R}^\dd$, for $k=1 \ldots \lambda$  (line \ref{CMAsampling}): 
	
	\begin{equation}
  \vc{x}^t_k = {\mathcal N}  \hspace{-0.13em}\left({\vc{m}^t,{\sigma^t}^2 {\C}^t}\right) = \vc{m}^t + \sigma^t {\mathcal N}  \hspace{-0.13em}\left({\ma{0},{\C}^t}\right),
  \end{equation}
	
	where the mean $\vc{m}^t \in \mathbb{R}^\dd$ of the distribution can be interpreted as the current estimate of the optimum of function $f$,  $\C^t \in \R^{n \times n}$ 
is a (positive definite) covariance matrix and $\sigma^t$ is a mutation step-size. These $\lambda$ solutions are evaluated according to $f$ (line \ref{CMAGenerateEnd}).
	The new mean $\vc{m}^{t+1}$ of the distribution is computed as a \textit{weighted sum} 
	of the best $\mu$ individuals out of the $\lambda$ ones (line \ref{CMAComputeNewMean}).
	Weights $w_1 \ldots w_\mu$ are used to control the impact of selected individuals, with usually higher weights for top ranked individuals (line \ref{CMAEScmaGiven}).
	
	The adaptation of the step-size $\sigma^t$, inherited from the Cumulative Step-Size Adaptation Evolution Strategy (CSA-ES \cite{1996HansenCMAES}), is controlled by the evolution path $\vc{p}_{\sigma}^{t+1}$.
	Successful mutation steps $\frac{\vc{m}^{t+1}-\vc{m}^{t}}{\sigma^{t}}$ (line \ref{CMASigmaPathUpdate}) are tracked in the sampling space, i.e., in the isotropic coordinate system defined by the eigenvectors of the covariance matrix $\C^t$. To update the evolution path $\vc{p}_{\sigma}^{t+1}$, i) a decay/relaxation factor $c_{\sigma}$ is used to decrease the importance of previous steps; ii) the step-size is increased if the length of the evolution path $\vc{p}_{\sigma}^{t+1}$ is longer than 
	the expected length of the evolution path under random selection $\mathbb{E} \left\| \NormOI \right\|$; iii) otherwise it is decreased  (line \ref{CMAStepSizeUpdate}). Expectation of $\left\| \NormOI \right\|$ is approximated by $\sqrt{n} (1 - \frac{1}{4 n} + \frac{1}{21 n^2} )$.
	A damping parameter $d_{\sigma}$ controls the change of the step-size.
	
	The covariance matrix update consists of two parts (line \ref{CMAupdate}): a \textit{rank-one update} \cite{2001HansenCMA} and a \textit{ rank-$\mu$ update} \cite{2003HansenCMA}.	The rank-one update computes the evolution path $\vc{p}_c^{t+1}$ of successful moves of the mean $\frac{\vc{m}^{t+1}-\vc{m}^{t}}{\sigma^{t}}$ 
	of the mutation distribution in the given coordinate system (line \ref{CMAEvoPathUpdate}), 
	along the same lines as the evolution path $\vc{p}_{\sigma}^{t+1}$ of the step-size. 
	To stall the update of $\vc{p}_c^{t+1}$ when $\sigma$ increases rapidly, a $h_{\sigma}$ trigger is used (line \ref{CMAhsigma}).

\begin{algorithm}[tb!]
\caption{The ($\mu/\mu_{w},\lambda$)-CMA-ES}
\label{CMAdefault}
\begin{algorithmic}[1]
\STATE{\textbf{given} $n \in \mathbb{N}_+$, $\lambda = 4 + \lfloor 3 \mstr{ln} \, n  \rfloor $, $\mu =  \lfloor \lambda/2   \rfloor $, 
											$w_i = \frac{ \mstr{ln}(\mu + \frac{1}{2}) - \mstr{ln}\,i}{ \sum^{\mu}_{j=1}(\mstr{ln}(\mu + \frac{1}{2})-\mstr{ln}\,j)} \; \mstr{for} \; i=1 \ldots \mu$,
											$\mu_w = \frac{1}{\sum^{\mu}_{i=1} w^2_i}$,
											$c_{\sigma} = \frac{\mu_w + 2}{n+\mu_w+3}$,      $d_{\sigma} = 1 + c_{\sigma} +2 \, \mstr{max}(0,\sqrt{ \frac{\mu_w - 1}{n+1}}-1)$,
											$c_c = \frac{4}{n+4}$, $c_1 = \frac{2}{(n+1.3)^2 +\mu_w}$, $c_{\mu} = \frac{2 \, (\mu_w -2 + 1/{\mu_w})}{(n+2)^2+\mu_w}$}
											 \label{CMAEScmaGiven}
\STATE{\textbf{initialize} $\vc{m}^{t=0} \in \R^{\dd}, \sigma^{t=0} > 0, \vc{p}^{t=0}_{\sigma} = \ma{0}, \vc{p}^{t=0}_{c} = \ma{0}, \C^{t=0} = \Id, t \leftarrow 0 $}
\REPEAT
  \FOR{$k = 1,\ldots,\lambda$} \label{CMAGenerateBegin}
			\STATE{ $\vc{x}_k = \vc{m}^t + \sigma^{t}  {{\mathcal N}  \hspace{-0.13em}\left({\ma{0},\C^{t}\,}\right)} $} \label{CMAsampling}
			\STATE{ $\vc{f}_k = f(\vc{x}_k)$} \label{CMAGenerateEnd}
  \ENDFOR
	\STATE{ $ \vc{m}^{t+1} = \sum_{i=1}^{\mu} \vc{w}_i \vc{x}_{i:\lambda} \;$} // the symbol $i:\lambda$ denotes $i$-th best individual on $f$ \label{CMAComputeNewMean}
	\STATE{ $ \vc{p}^{t+1}_{\sigma} = (1 - c_{\sigma}) \vc{p}^{t}_{\sigma} + \sqrt{c_{\sigma}(2-c_{\sigma})} \sqrt{\mu_w} {\C^t}^{-\frac{1}{2}} \frac{\vc{m}^{t+1}-\vc{m}^{t}}{\sigma^{t}} $} \label{CMASigmaPathUpdate}
	\STATE{ $ h_{\sigma} = \ONE_{ \left\| p^{t+1}_{\sigma} \right\| < \sqrt{1 - (1-c_{\sigma})^{2(t+1)}}(1.4 + \frac{2}{n+1}) \, \mathbb{E} \left\| \NormOI \right\|  } $} \label{CMAhsigma}
	\STATE{ $ \vc{p}^{t+1}_{c} = (1 - c_{c}) \vc{p}^{t}_{c} + h_{\sigma} \sqrt{c_{c}(2-c_{c})} \sqrt{\mu_w} \frac{\vc{m}^{t+1}-\vc{m}^{t}}{\sigma^{t}} $} \label{CMAEvoPathUpdate}
	\STATE{ $ \C_{\mu} = \sum^{\mu}_{i=1} w_i \frac{\x_{i:\lambda} - \m^t}{\sigma^t} \times \frac{(\x_{i:\lambda} - \m^t)^T}{\sigma^t}$ }  \label{CMAplusCov}
	\STATE{ $ \C^{t+1} = (1 - c_1 - c_{\mu}) \C^t	+ 
						c_1 \underbrace{\vc{p}^{t+1}_c {\vc{p}^{t+1}_c}^T}_{\mstr{\tiny rank-one\,update}} + 
						c_{\mu} \hspace{-1.9em} \underbrace{\C_{\mu}}_{\mstr{rank-\mu \,update}} $}
						\label{CMAupdate}
	\STATE{ $ \sigma^{t+1} = \sigma^{t} \mstr{exp}
	          ( \frac{c_{\sigma}}{d_{\sigma}} (  \frac{\left\| \vc{p}^{t+1}_{\sigma} \right\|}{ \mathbb{E} \left\| \NormOI \right\| } - 1  )) $} \label{CMAStepSizeUpdate}
  \STATE{ $ t = t + 1$} \label{endupdate}
\UNTIL{ \textit{stopping criterion is met} }
\end{algorithmic}
\end{algorithm}

The rank-$\mu$ update computes a covariance matrix $\C_{\mu}$ as a weighted sum of the covariances of successful steps of the best $\mu$ individuals (line \ref{CMAplusCov}).
	Covariance matrix $\C$ itself is replaced by a weighted sum of the rank-one (weight $c_1$ \cite{2001HansenCMA}) and rank-$\mu$ (weight $c_\mu$  \cite{2003HansenCMA}) updates, with $c_1$ and $c_\mu$ positive and $c_1 + c_{\mu}  \leq 1$.
	
	While the optimal parameterization of CMA-ES remains an open problem, the default parameterization is found quite robust on  noiseless unimodal functions \cite{2003HansenCMA}, which explains the popularity of CMA-ES. 



\section{The \Free-CMA-ES}
\label{section:Approach}

The proposed \Free-CMA-ES approach is based on the intuition that the optimal hyper-parameters of CMA-ES at time $t$ should favor the generation of the best individuals at time $t$, under the (strong) assumption that an optimal parameterization and performance of CMA-ES in each time $t$ will lead to the overall optimal performance. 

Formally, this intuition leads to the following procedure. Let $\theta^t_f$ denote the hyper-parameter vector used for the optimization of objective $f$ at time $t$ (CMA-ES stores its state variables and internal parameters of iteration $t$ in $\theta^t$ and the '.'-notation is used to access them). At time $t+1$, the best individuals generated according to $\theta^t_f$ are known to be the top-ranked individuals $\vc{x}^t_{1:\lambda} \ldots \vc{x}^t_{\mu:\lambda} $, where $\vc{x}^t_{i:\lambda}$ stands for the $i$-th best individual w.r.t. $f$. Hyper-parameter vector $\theta^t_f$ would thus have been all the better, if it had maximized the probability of generating these top individuals. 

\def\aux{{auxiliary}}
\def\prim{{primary}}
Along this line, the optimization of $\theta^t_f$ is conducted using a second CMA-ES algorithm, referred to as {\em \aux} CMA-ES as opposed to the one concerned with the optimization of $f$, referred to as {\em \prim} CMA-ES. The objective of the \aux\ CMA-ES is specified as follows: 
%

\textbf{Given}: hyper-parameter vector $\theta^i_f$ and points $(\vc{x}^i_{1:\lambda}, f(\vc{x}^i_{1:\lambda}))$
evaluated by \prim\ CMA-ES at steps $i=1,\ldots,t$ (noted as $\theta^{i+1}_{f}.f(\vc{x}_{1:\lambda})$ in Algorithm \ref{algoalgo}), 

\textbf{Find}: $\theta^{t,\ast}_f$ such that i) backtracking the \prim\ CMA-ES to its state at time $t-1$; ii) replacing $\theta^{t}_f$ by $\theta^{t,\ast}_f$, would maximize the likelihood of $\vc{x}^t_{i:\lambda}$ for $i = 1 \ldots \mu$. 


The \aux\ CMA-ES might thus tackle the maximization of $g_t(\theta)$ computed as the weighed log-likelihood of the top-ranked $\mu_{sel}$ individuals at time $t$:
	\begin{equation}
g_t(\theta) = \sum_{i=1}^{\mu_{sel}} w_{sel,i} \log \left (P(\vc{x}^t_{i:\lambda} | \theta^t_f = \theta \right)),
	\label{eq:lk}
  \end{equation}
where $w_{sel,i} \geq 0, \; i=1 \ldots \mu_{sel}, \sum_{i=1}^{\mu_{sel}} w_{sel,i} = 1,$ and by construction 
	\begin{equation}
  P(\vc{x}_{i}| \vc{m}^t, \vc{C}^t) = \frac{1}{\sqrt{{(2\pi)}^n |\vc{C}^t|}}{\exp{(-0.5 (\vc{m}^t - \vc{x}^t_i) {\vc{C}^t}^{-1} (\vc{m}^t - \vc{x}^t_i))}},
  \end{equation}
where $\vc{C} ^t$ is the covariance matrix multiplied by ${\sigma^t}^2$ and $|\vc{C}^t|$ is its determinant.

While the objective function for the \aux\ CMA-ES defined by Eq. (\ref{eq:lk}) is mathematically sound, it yields a difficult optimization problem; firstly the probabilities are scale-sensitive; secondly and overall, in a worst case scenario, a single good but unlikely solution may lead the optimization of $\theta^t_f$ astray.\niko{I don't buy the argument, given how the algorithm is implemented later on, in particular because the secondary CMA-ES is only run for a single iteration. On this note, I think the rank correlation as fitness has the problem to produce plateaus. }\ilya{I suggest to make the second argument a bit weaker (e.g., 'can lead to' instead of 'is observed to') because we have data to demonstrate it happening. Otherwise,+ we may be too confident (but without enough data) to kill the direction of the original likelihood while who knows how promising it is in long term}\niko{(i) Auxiliary CMA is rank-based, therefore the optimization cannot go astray in one iteration, it just cannot. That exactly is the beauty of rank-based algorithms. (You can use entirely random fitness values every second iteration and still optimize the ellipsoid function, about five times slower.) So, the argument/explanation is simply wrong, as far as I can see. (ii) what are these data? }
 
Therefore, another optimization objective $h_t(\theta)$ is defined for the \aux\ CMA-ES, where $h_t(\theta)$ measures the agreement on $\vc{x}^t_{i:\lambda}$ for $i=1 \ldots \mu$ between i) the order defined from $f$; ii) the order defined from their likelihood conditioned by $\theta^t_f = \theta$ (Algorithm \ref{alfunch}). Procedure \textit{ReproduceGenerationCMA} in Algorithm~\ref{alfunch} updates the strategy parameters described from line \ref{CMAComputeNewMean} to line \ref{endupdate} in Algorithm \ref{CMAdefault} using already evaluated solutions stored in $\theta^{t}_{f}.\vc{x}_{i:\lambda}$. Line 4 computes the Mahalanobis distance, division by step-size is not needed since only ranking will be considered in line 5 (decreasing order of Mahalanobis distances corresponds to increasing order of log-likelihoods). Line 6 computes a weighted sum of ranks of likelihoods of best individuals. 

	Finally, the overall scheme of \Free-CMA-ES (Algorithm \ref{algoalgo}) involves two interdependent CMA-ES optimization algorithms, where the \prim\ CMA-ES is concerned with optimizing objective $f$, and
the \aux\ CMA-ES is concerned with optimizing objective $h_t$, that is, optimizing the hyper-parameters of the \prim\ CMA-ES\footnote{This scheme is actually inspired from the one proposed for surrogate-assisted optimization \cite{2012LoshchilovSAACMGECCO}, where the \aux\ CMA-ES was in charge of optimizing the surrogate learning hyper-parameters.}.
Note that \Free-CMA-ES is not {\em per se} a ``more parameterless`` algorithm than CMA-ES, in the sense that the user is still invited to modify the population size $\lambda$. The main purpose of \Free-CMA-ES is to achieve the online adaptation of the other CMA-ES hyper-parameters. 
	
	Specifically, while the \prim\ CMA-ES optimizes $f(\vc{x})$ (line \ref{oneiteronF}), the \aux\ CMA-ES maximizes $h_t(\theta)$ (line \ref{oneiteronH}) by sampling and evaluating $\lambda_{h}$ variants of $\theta^t_f$. The updated mean of the \aux\ CMA-ES in the hyper-parameter space is used as a local estimate of the optimal hyper-parameter vector for the \prim\ CMA-ES. Note that the \aux\ CMA-ES achieves a single iteration in the hyper-parameter space of the \prim\ CMA-ES, with two motivations: limiting the 
computational cost of \Free-CMA-ES (which scales as $\lambda_{h}$ times the time complexity of the CMA-ES), and preventing $\theta^t_f$ from overfitting the current sample $\vc{x}^t_{i:\lambda}, i = 1 \ldots \mu$.

\def\thetaf{{\mbox{$\theta_{f}$}}}	
	\begin{algorithm}[t!]
\caption{The \Free-CMA-ES}
\label{algoalgo}
\begin{algorithmic}[1]
\STATE{$ t \leftarrow 1$} 
\STATE{ $\theta^t_{f}$ $\leftarrow$ InitializationCMA() }	\COMMENT{	\prim\ CMA-ES aimed at optimizing $f$ }	\label{algoinitf}
\STATE{ $\theta^t_{h}$ $\leftarrow$ InitializationCMA() }	\COMMENT{	\aux\ CMA-ES aimed at optimizing $h_t$  }	\label{algoinith}
\STATE{ fill $\theta^t_{f}$ with corresponding parameters stored in mean of distribution $ \theta^{t}_{h}.m$}
		\STATE{ $\theta^{t+1}_{f}$ $\leftarrow$ GenerationCMA($f$, $\theta^t_{f}$) }
		\STATE{ $ t \leftarrow t + 1$ }
\REPEAT 
	\STATE{ $ \theta^{t+1}_{f} \leftarrow $GenerationCMA$(f, \theta^t_{f})$ } \label{oneiteronF}
	\STATE{ $ \theta^{t+1}_{h} \leftarrow $GenerationCMA$(h_t,\theta^t_{h}$) } \label{oneiteronH}
	\STATE{ fill $\theta^{t+1}_{f}$ with corresponding parameters stored in mean of distribution $ \theta^{t+1}_{h}.m$}
	\STATE{$ t \leftarrow t + 1$} 
\UNTIL{stopping criterion is met }
\end{algorithmic}
\end{algorithm}

\def\mus{{\mbox{$\mu$}}}	

\begin{algorithm}[ht!]
\label{alfunch}
\caption{Objective function $h_t(\theta)$}
\begin{algorithmic}[1]
\STATE{ \textbf{Input}: $\theta$, $\theta^{t+1}_{f}$,$\theta^{t-1}_{f}$, $\theta^{t}_{f}$, \mus, $w_{sel,i}$ for $i=1,\ldots,\mus$}    
\STATE{ $\theta^{'t-1}_{f} \leftarrow \theta$}
\STATE{ $ \theta^{'t}_{f} \leftarrow $ReproduceGenerationCMA$(f, \theta^{'t-1}_{f})$  using already evaluated $\theta^{t}_{f}.\vc{x}_{i:\lambda}$ }
\STATE{ $ d_{i} \leftarrow \left\| \theta^{'t}_{f}.\sqrt{C^{-1}} \cdot (\theta^{t+1}_{f}.\vc{x}^t_{i} - \theta^{'t}_{f}.m) \right\|;$ for $i=1,\ldots,\theta^{t+1}_{f}.\lambda$ }
\STATE{ $ p_{i} \leftarrow$ rank of $d_i, i = 1\ldots \lambda$ sorted in decreasing order}
\STATE{ $ h(\theta) \leftarrow  \sum_{i=1}^{\mus} w_{sel,i} p_{i:\lambda} $} \COMMENT{	$i:\lambda$ denotes the rank of $\theta^{t+1}.\vc{x}_{i}$  }
\STATE{ \textbf{Output}: $h(\theta)$ }
\end{algorithmic}
\end{algorithm}

\begin{figure}[!h]
	\includegraphics[scale=0.372]{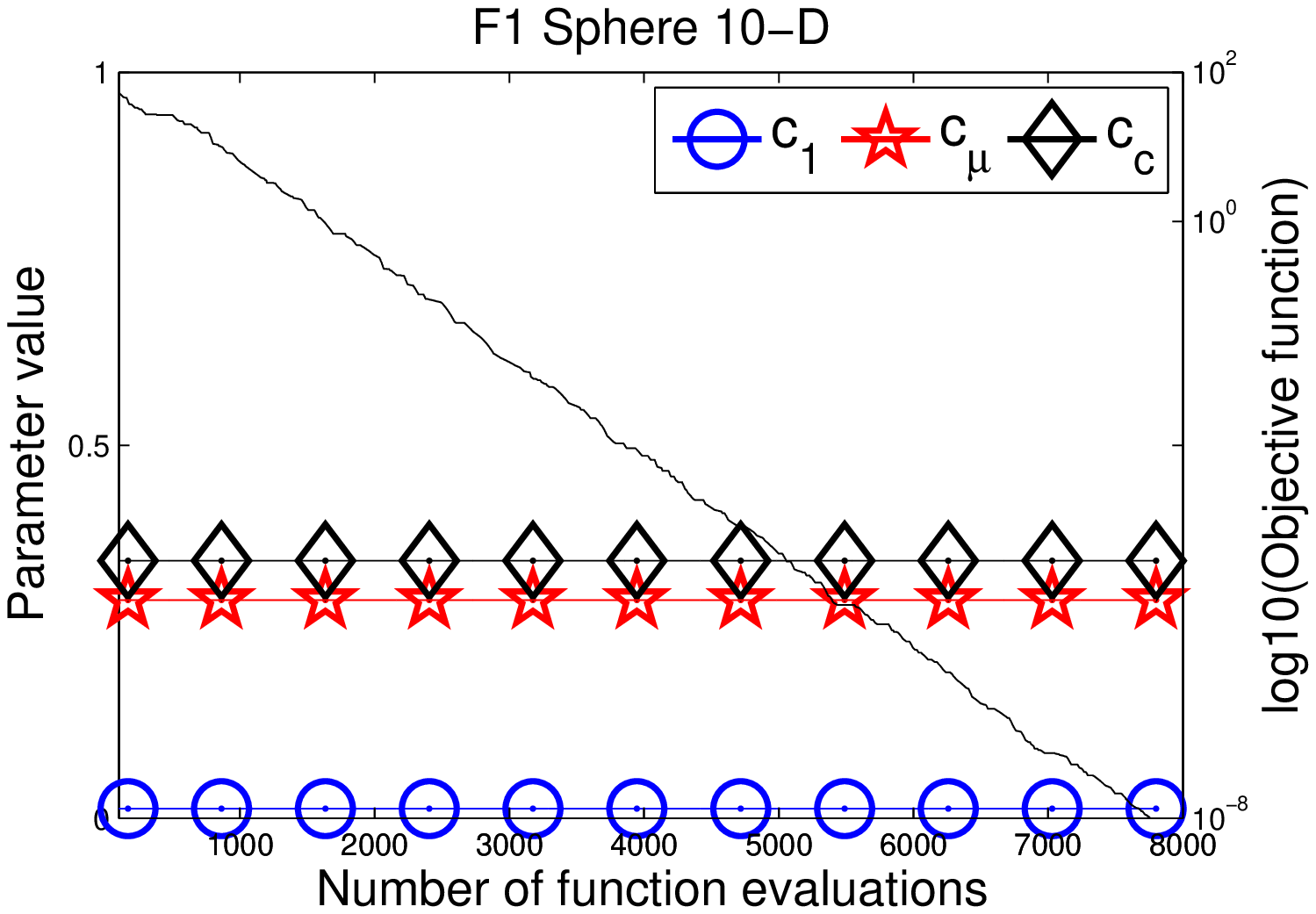}
  \includegraphics[scale=0.372]{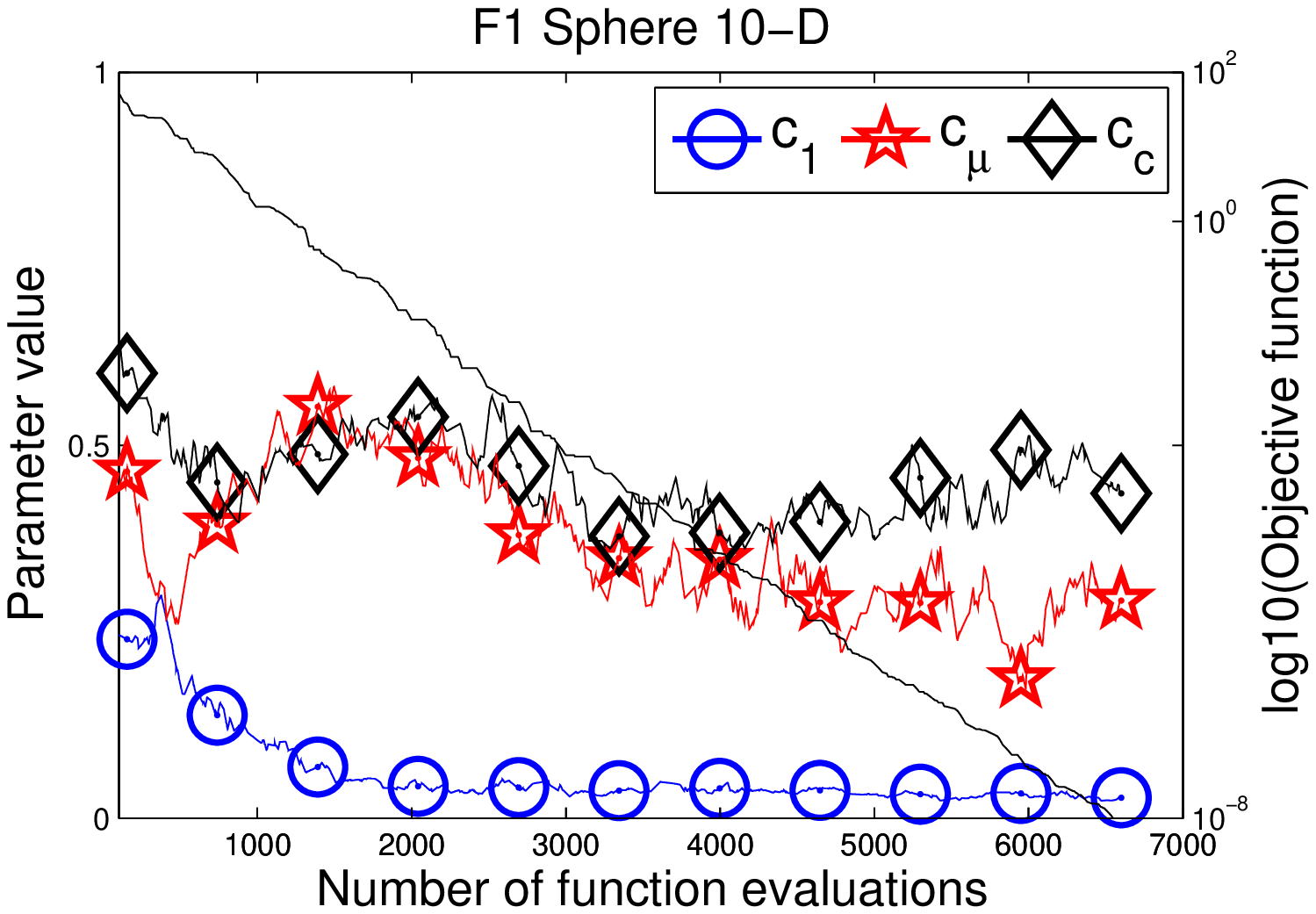} 	\\
	\includegraphics[scale=0.372]{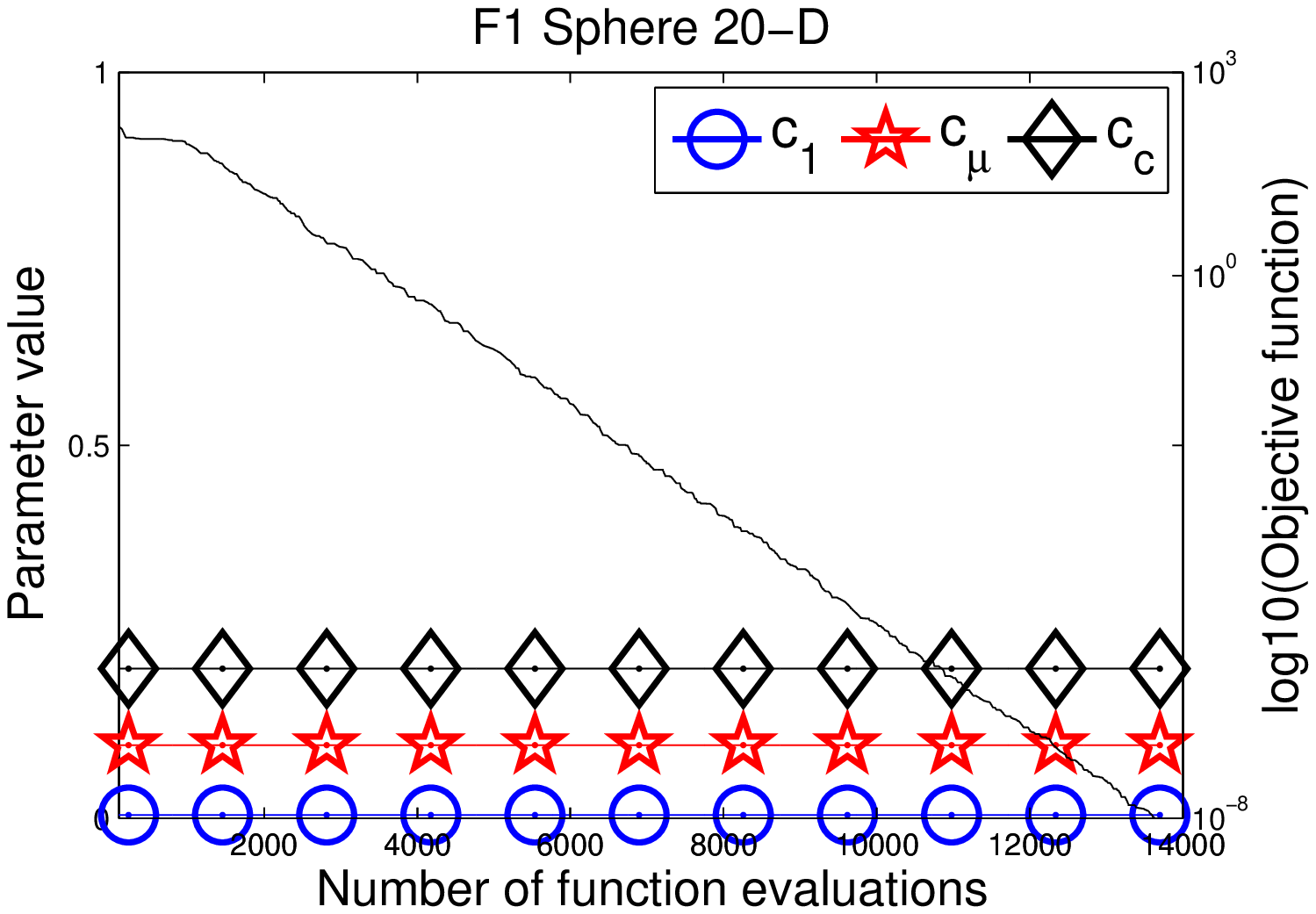}
  \includegraphics[scale=0.372]{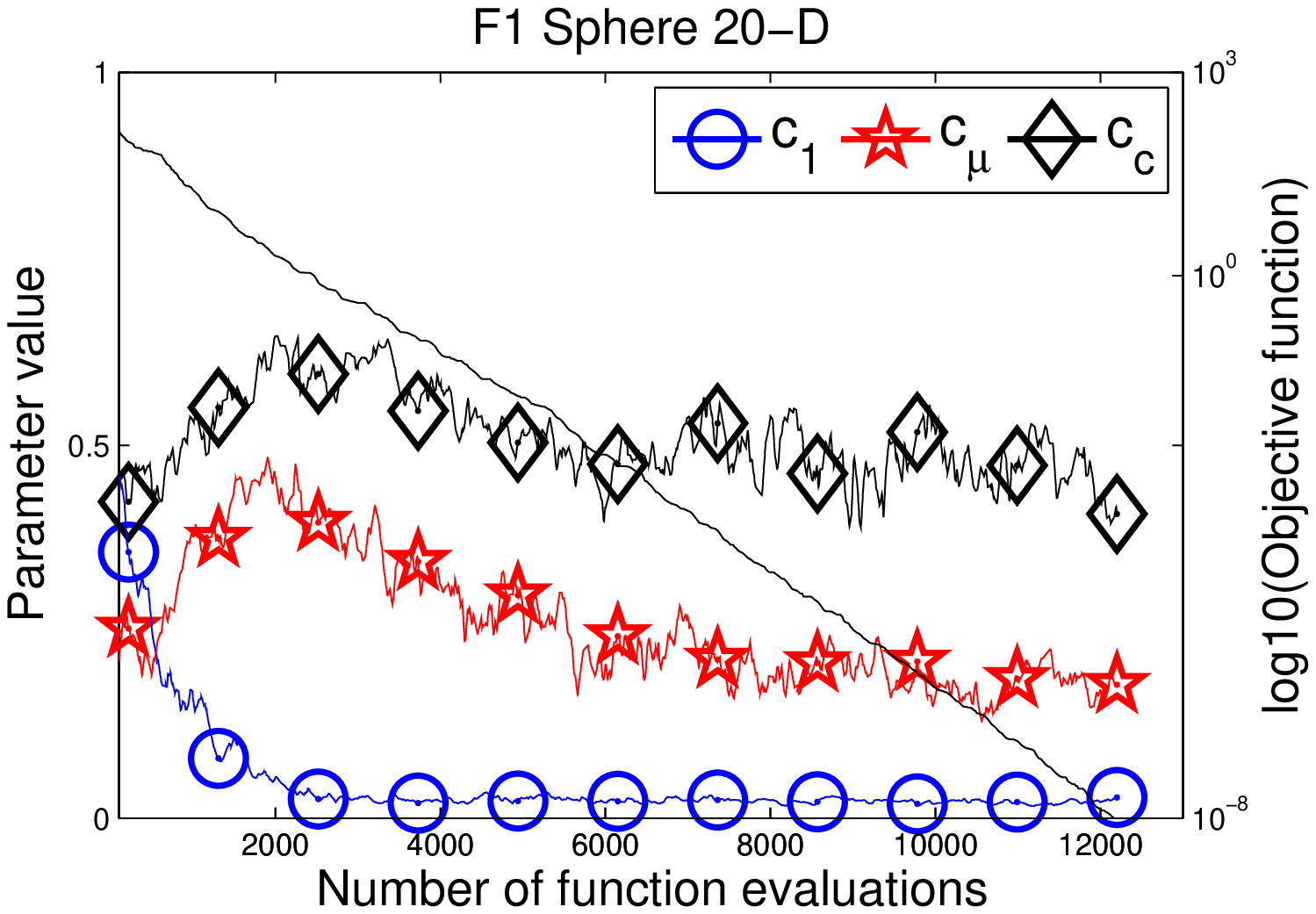}	\\
	\includegraphics[scale=0.372]{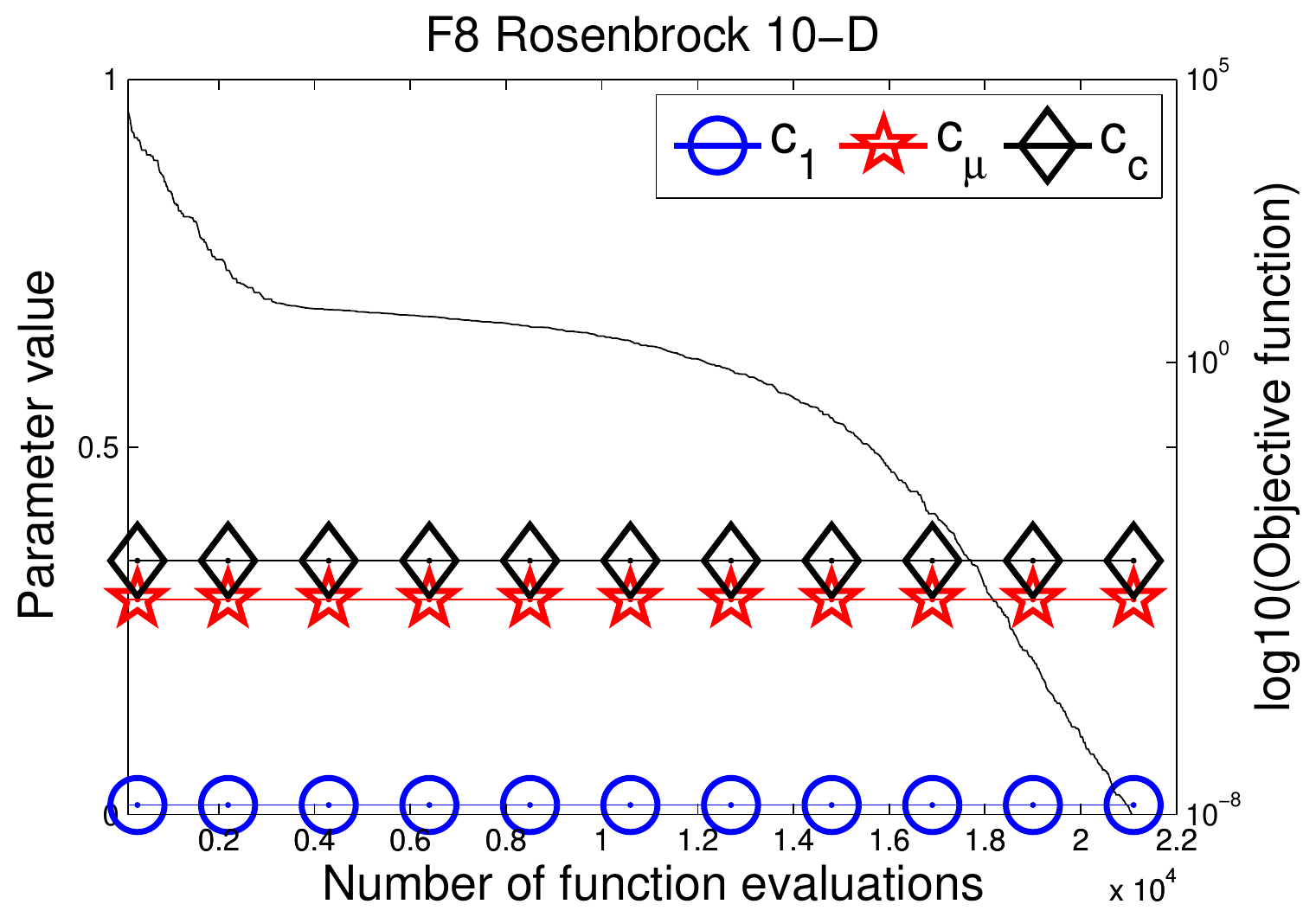}
  \includegraphics[scale=0.372]{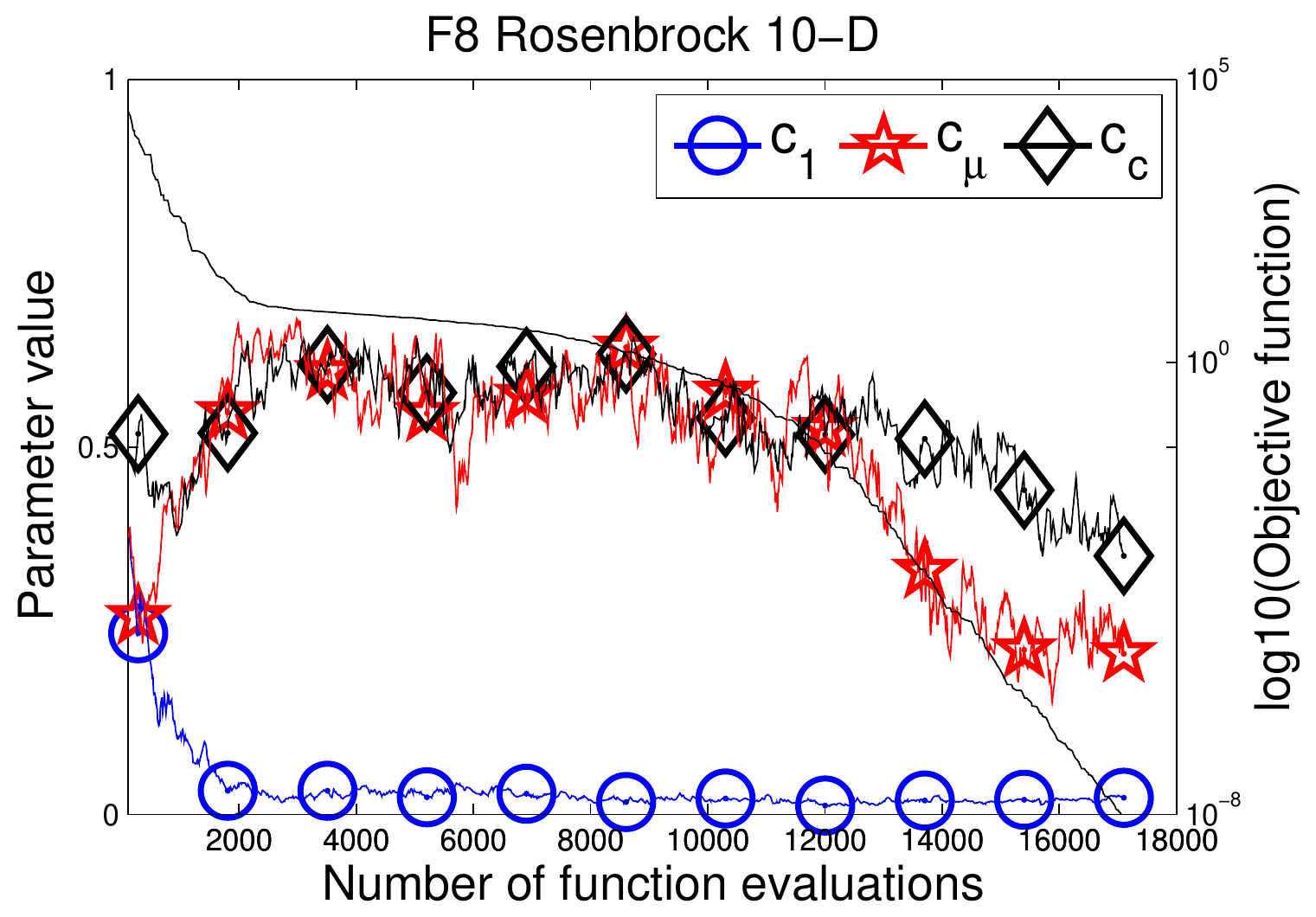} 	\\
	\includegraphics[scale=0.372]{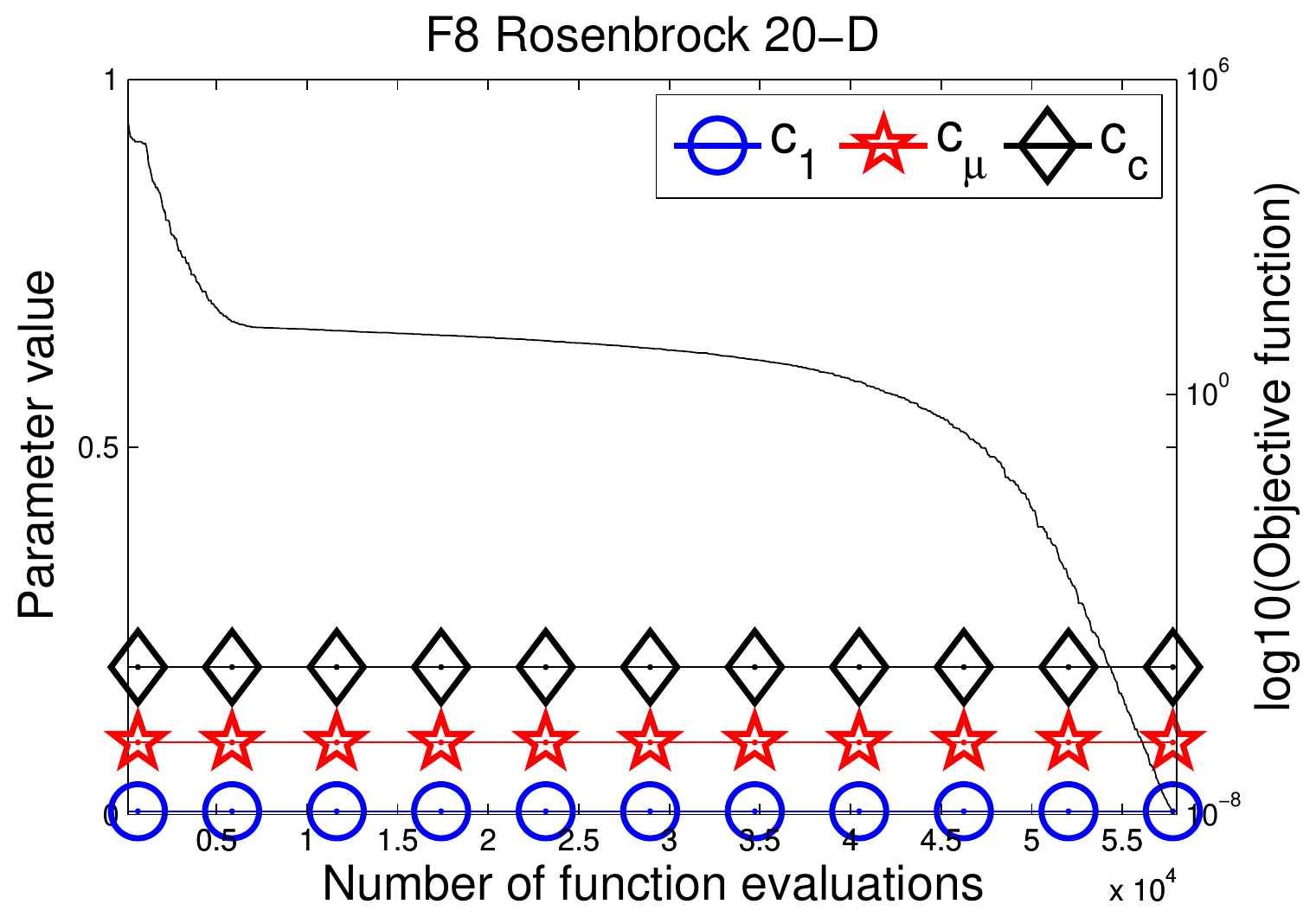}
  \includegraphics[scale=0.372]{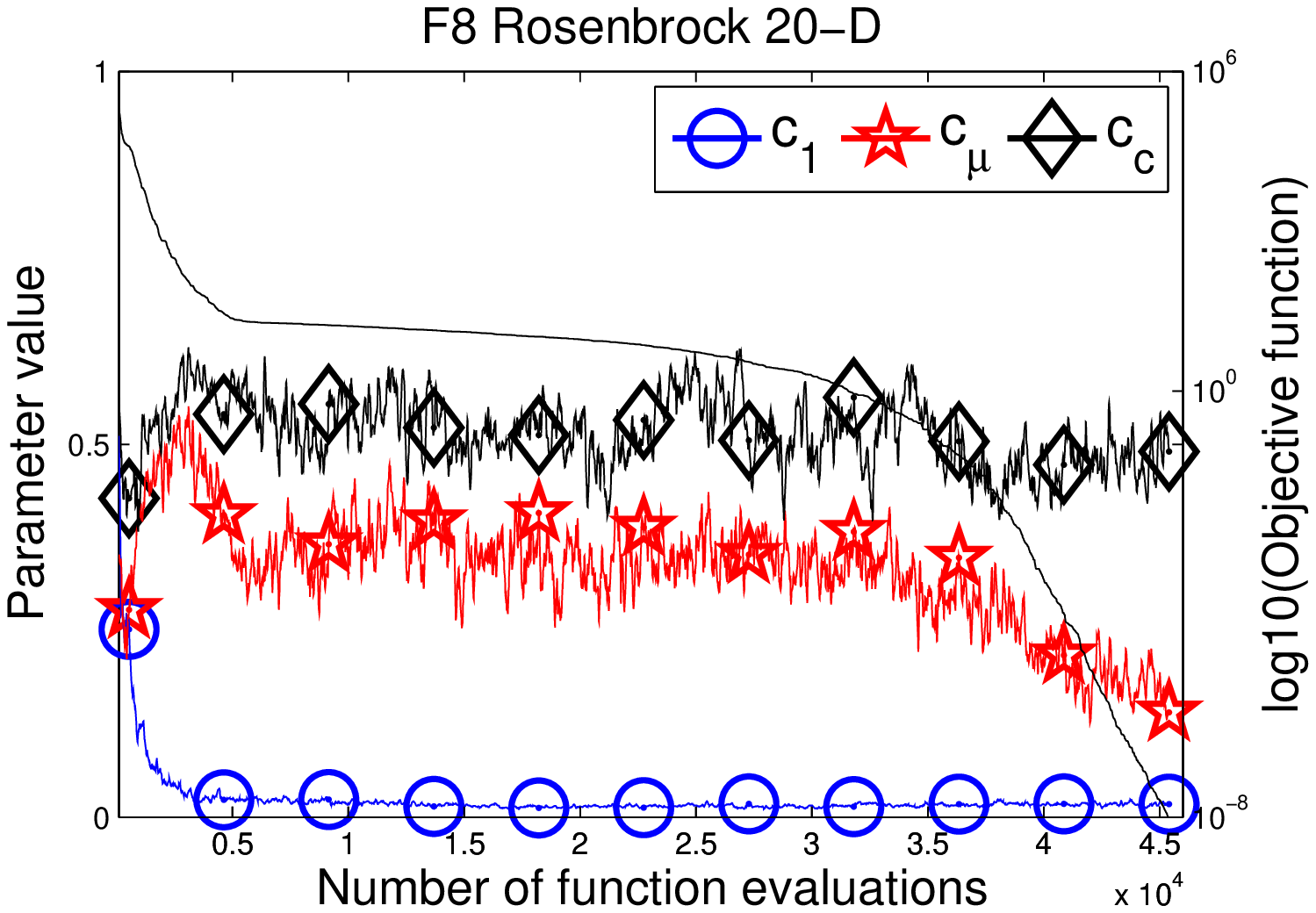}	\\
\caption{\label{fig1} Evolution of learning rates $c_1$, $c_{\mu}$, $c_c$ (lines with markers, left y-axis) and log10(objective function) (plain line, right y-axis) of CMA-ES (left column) and \Free-CMA-ES (right column) on 10- and 20-dimensional Sphere and Rosenbrock functions from \cite{2010HansenBBOBsetup}. The medians of 15 runs are shown.}
\end{figure}

\begin{figure}[!h]
	\includegraphics[scale=0.372]{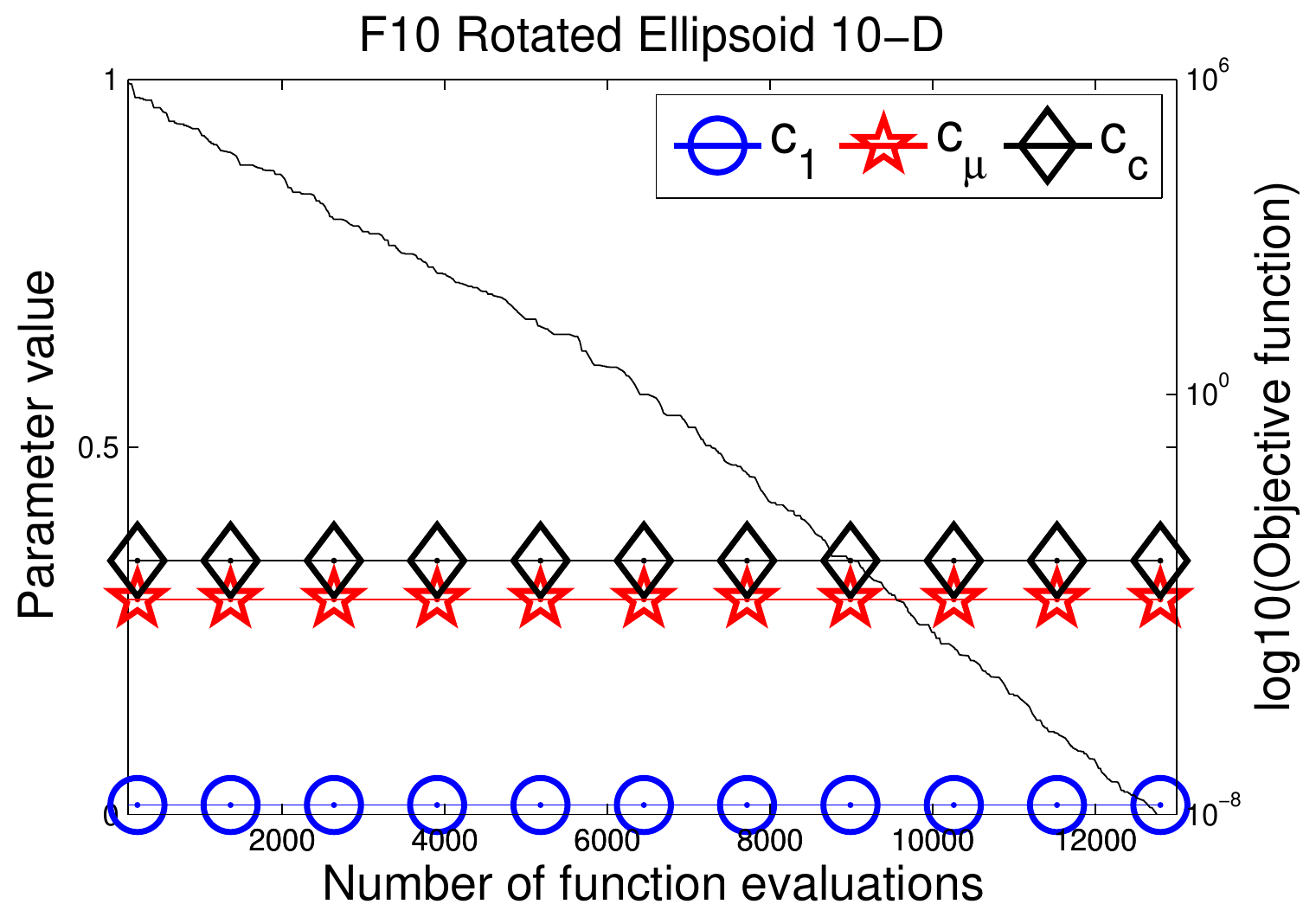}
  \includegraphics[scale=0.372]{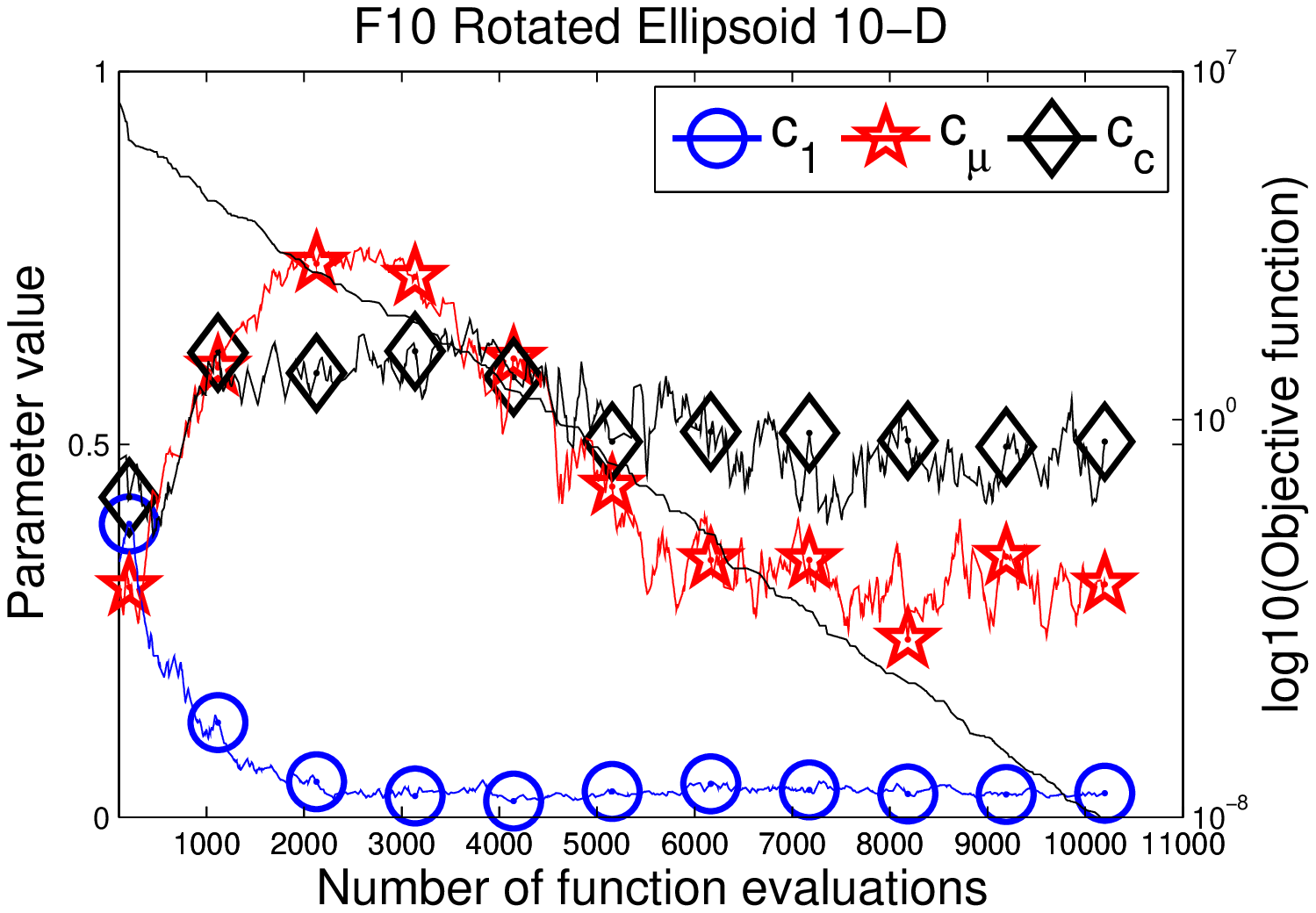} 	\\
	\includegraphics[scale=0.372]{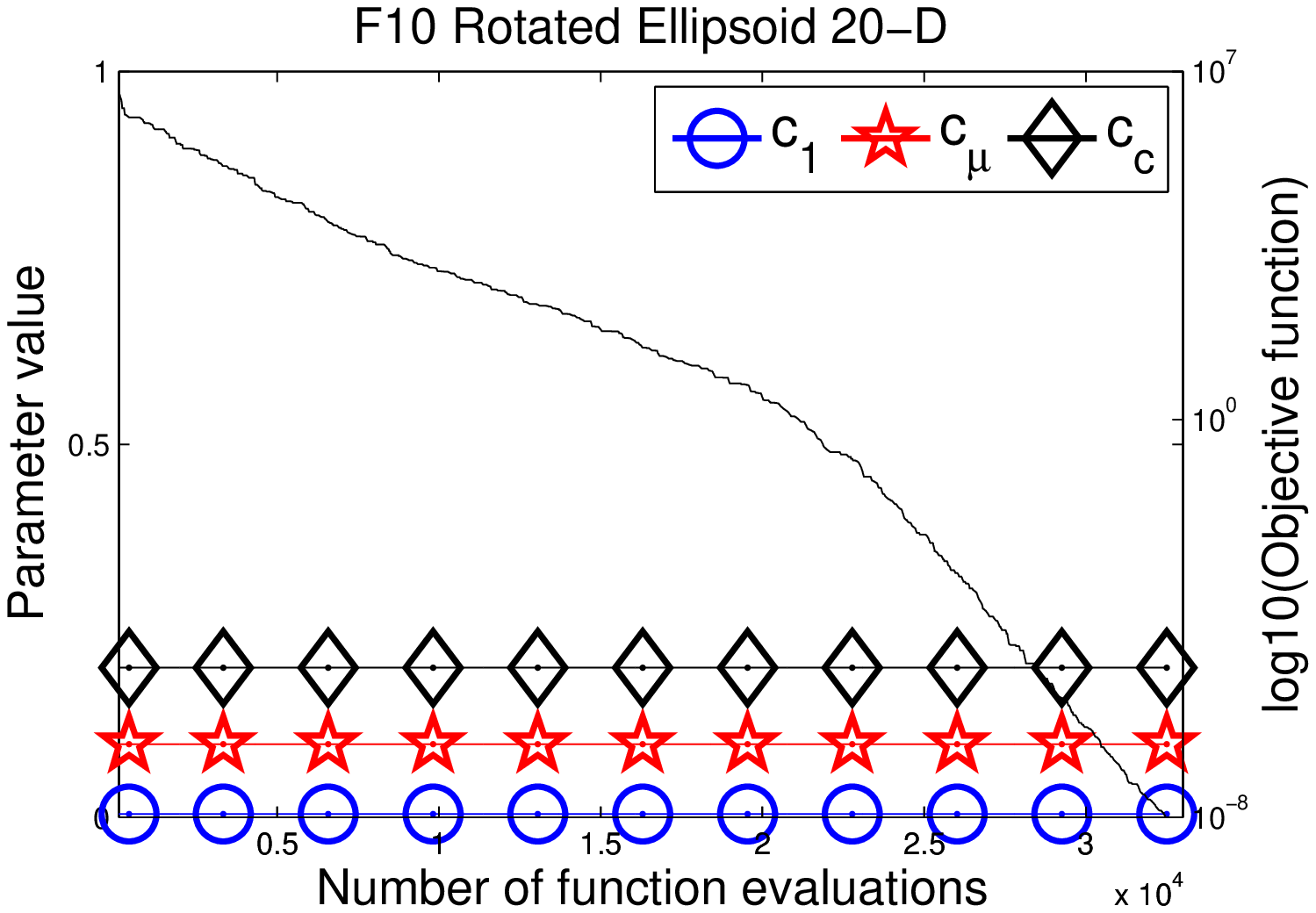}
  \includegraphics[scale=0.372]{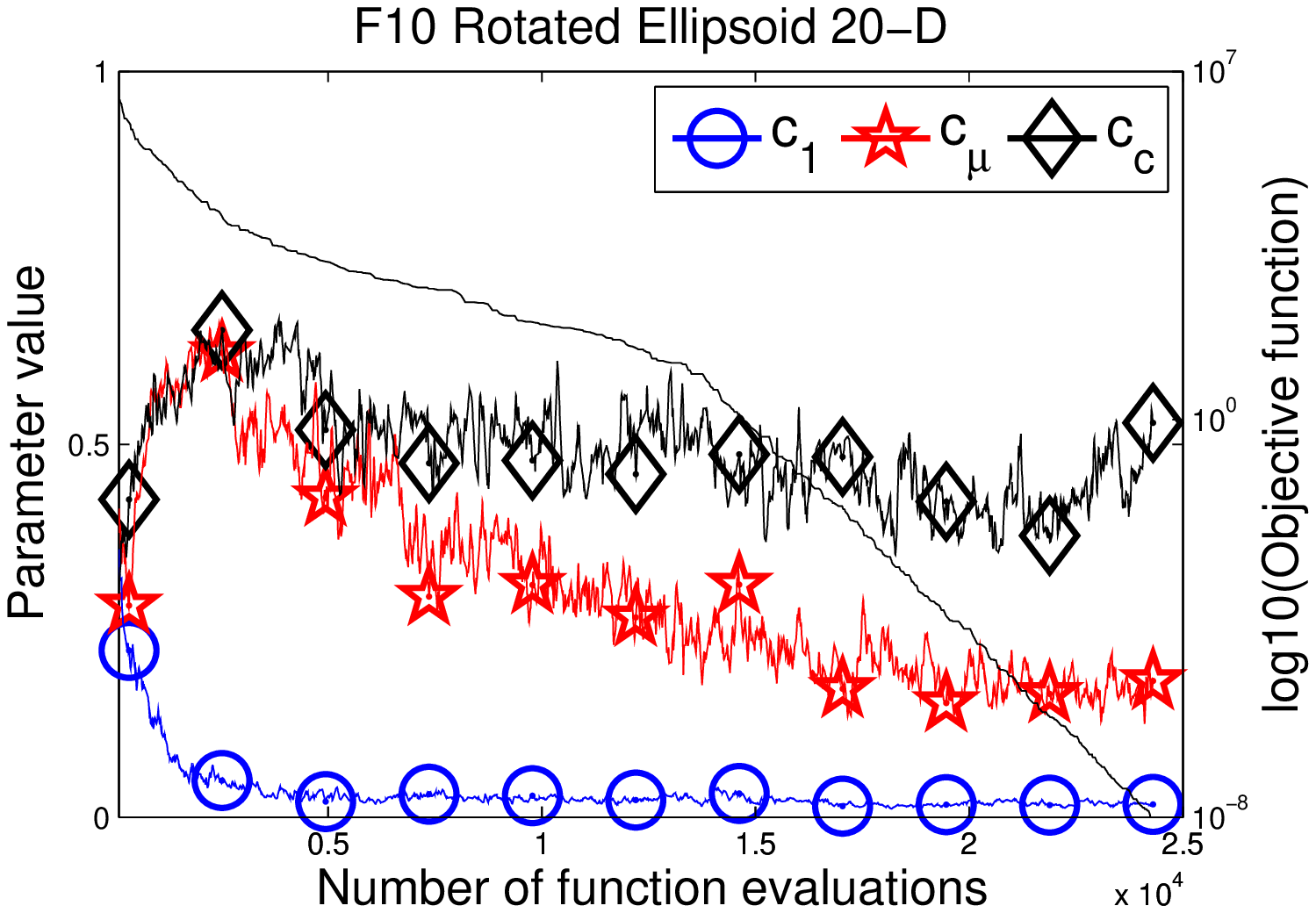}	\\
	\includegraphics[scale=0.372]{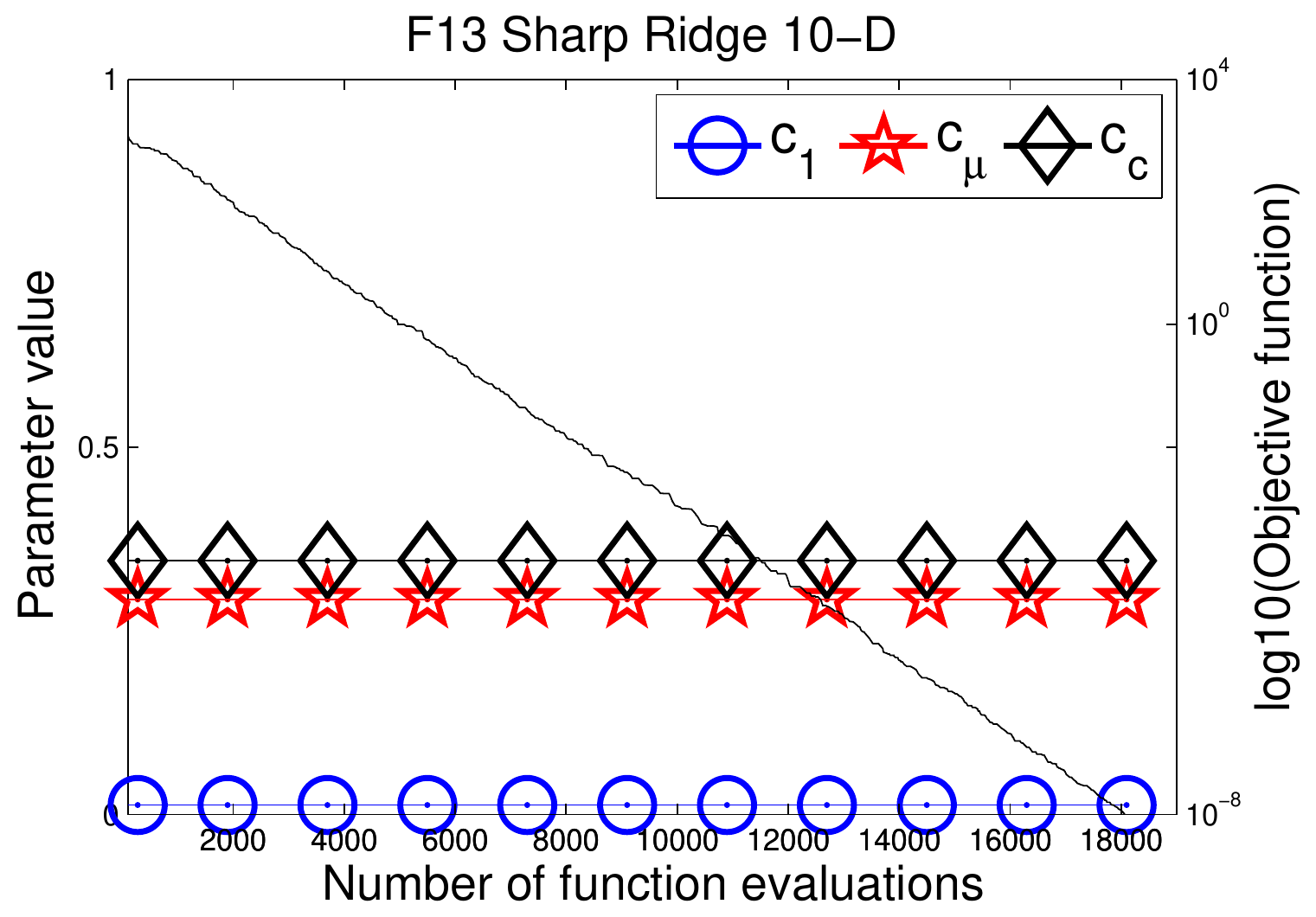}
  \includegraphics[scale=0.372]{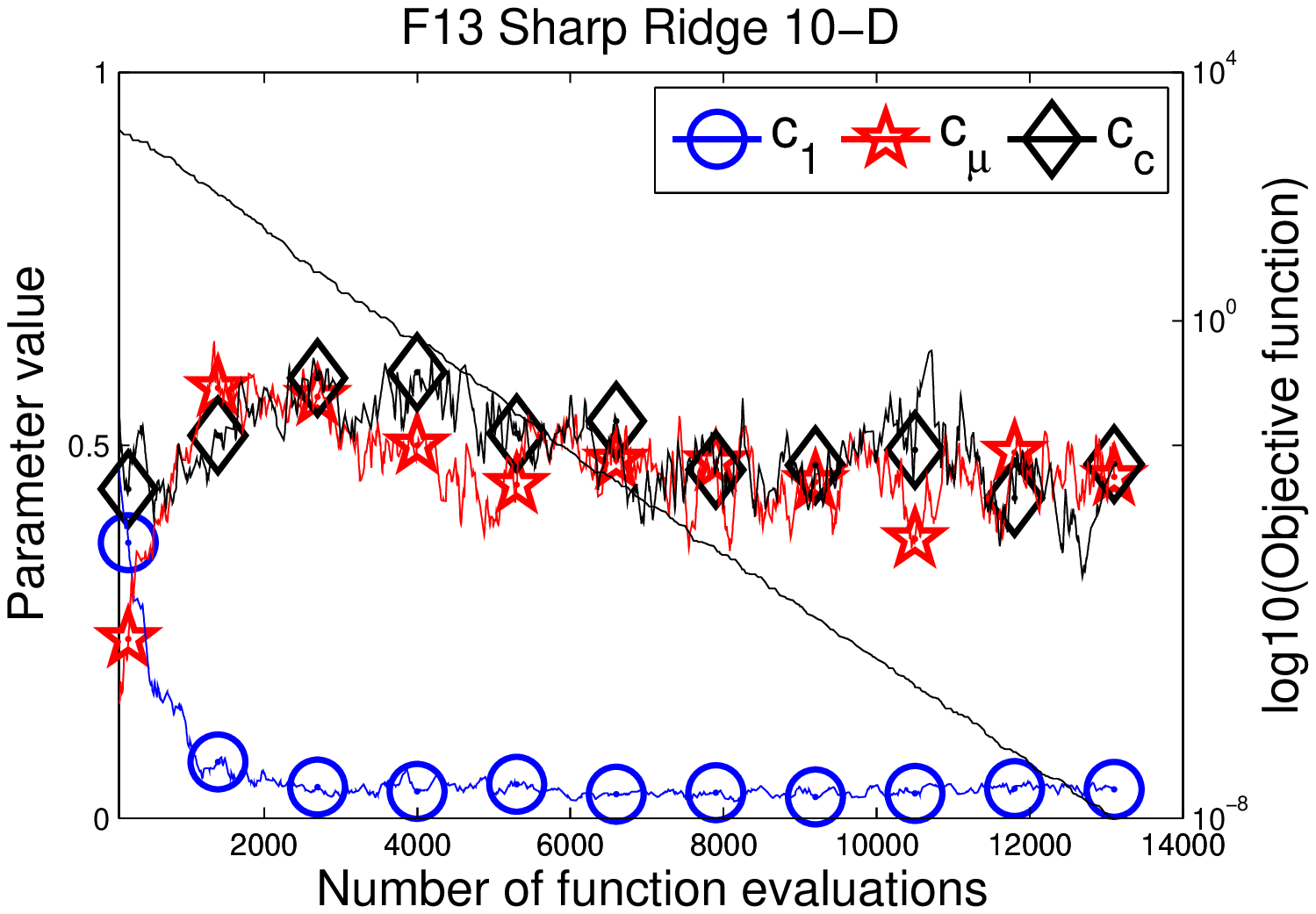} 	\\
	\includegraphics[scale=0.372]{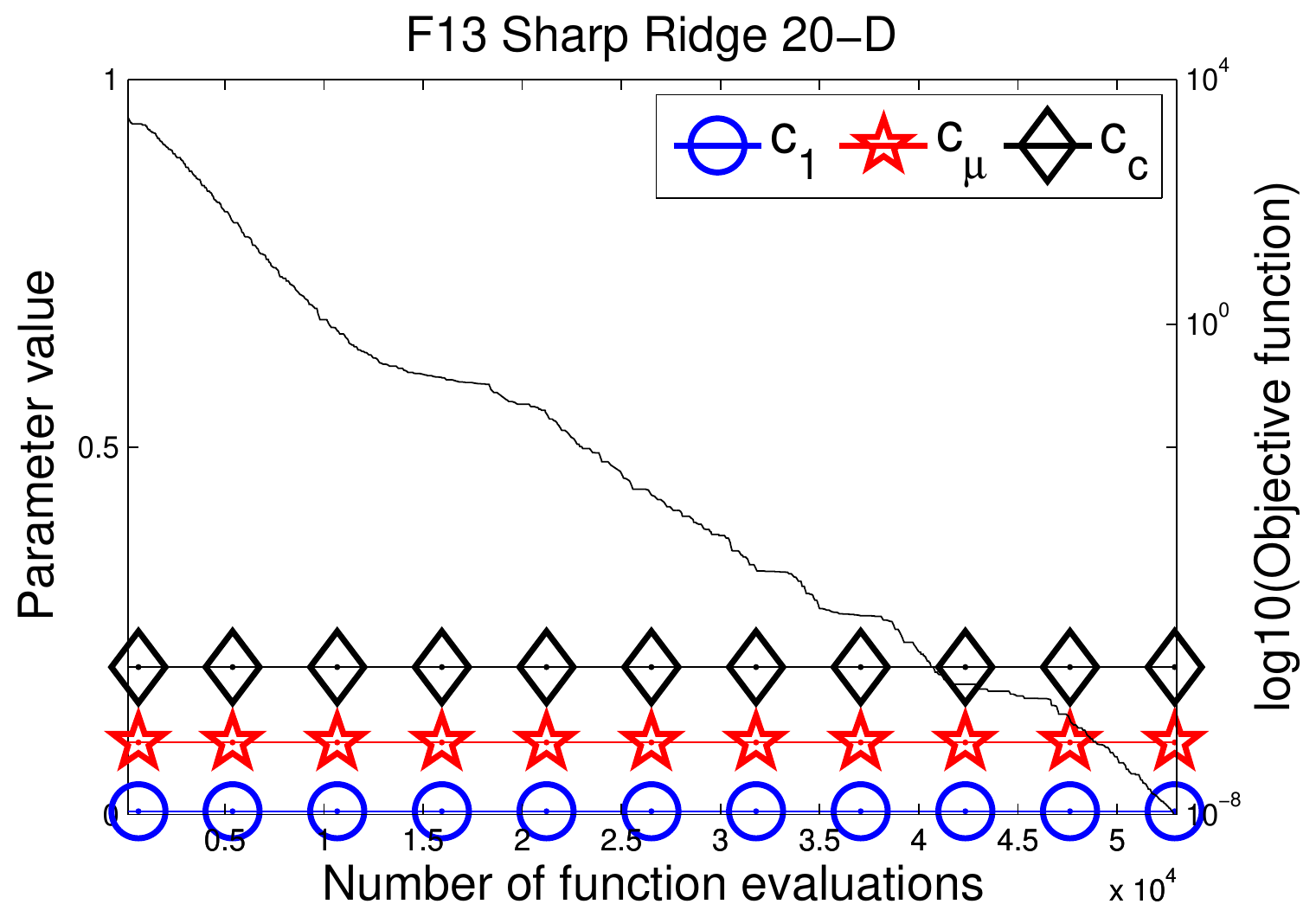}
  \includegraphics[scale=0.372]{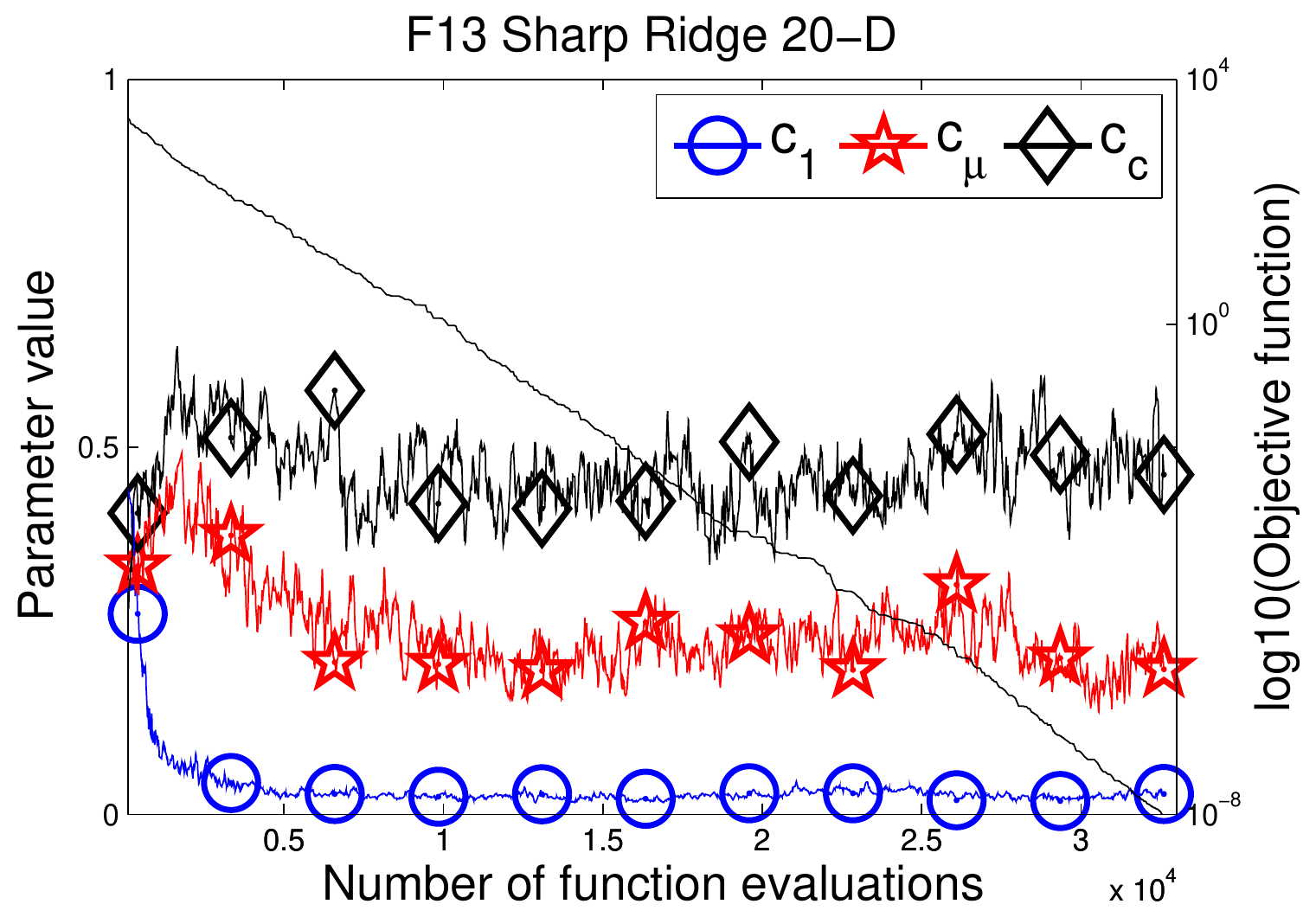}	\\
\caption{\label{fig2} Evolution of learning rates $c_1$, $c_{\mu}$, $c_c$ (lines with markers, left y-axis) and log10(objective function) (plain line, right y-axis) of CMA-ES (left column) and \Free-CMA-ES (right column) on 10- and 20-dimensional Rotated Ellipsoid and Sharp Ridge functions from \cite{2010HansenBBOBsetup}. The medians of 15 runs are shown.}
\end{figure}	

\section{Experimental Validation}\label{sectio:experiments}
The experimental validation of \Free-CMA-ES investigates the performance of the
algorithm comparatively to CMA-ES on the BBOB noiseless problems \cite{2010HansenBBOBsetup}. 
Both algorithms are launched in IPOP scenario of restarts when the CMA-ES is restarted 
with doubled population size once stopping criteria \cite{2009HansenBIPOP} are met\footnote{For the sake of reproducibility, the source code is available at \\ \textcolor[rgb]{0,0,1}{\url{https://sites.google.com/site/selfcmappsn/}}}.
 The population size $\lambda$ is chosen to be 100 for both CMA-ES and \Free-CMA-ES. 
We choose this value (about 10 times larger than the default one, see the default parameters of CMA-ES in Algorithm \ref{CMAdefault}) to investigate how sub-optimal the other CMA-ES hyper-parameters, derived from $\lambda$, are in such a case, and whether \Free-CMA-ES can 
recover from this sub-optimality.

The \aux\ CMA-ES is concerned with optimizing hyper-parameters $c_1$, $c_{\mu}$ and $c_c$ (Algorithm \ref{CMAdefault}), responsible for the adaptation of the covariance matrix of the 
\prim\ CMA-ES. These parameters range in $[0, .9]$ subject to $0 \leq c_1 + c_{\mu} \leq 0.9$; 
the constraint is meant to enforce a feasible  $\vc{C}$  update for the \prim\ CMA-ES (the decay factor of $\vc{C}$ should be in $[0,1]$). Infeasible hyper-parameter solutions get a very large penalty, multiplied by the sum of distances of infeasible hyper-parameters to the range of feasibility. 

We set $w_{sel,i}=1/\mu$ for $i=1,\ldots,\mus$ and $\mus=\left\lfloor \lambda/2\right\rfloor$ to 50. The internal computational complexity of \Free-CMA-ES thus is $\lambda_{h}=20$ times larger than the one of CMA-ES without lazy update (being reminded that the internal time complexity is usually negligible compared to the cost per objective function evaluation). 


\subsection{Results}

Figures \ref{fig1} and \ref{fig2} display the comparative performances of CMA-ES (left) and \Free-CMA-ES (right) on 10 and 20-dimensional Sphere, Rosenbrock, Rotated Ellipsoid and Sharp ridge functions from the noiseless BBOB testbed \cite{2010HansenBBOBsetup} (medians out of 15 runs).
Each plot shows the value of the hyper-parameters (left y-axis)
together with the objective function (in logarithmic scale, right y-axis). Hyper-parameters $c_1$, $c_{\mu}$ and $c_c$ are constant and set to their default values for CMA-ES while they are adapted along evolution for \Free-CMA-ES. 

In \Free-CMA-ES, the hyper-parameters are uniformly initialized in $[0,0.9]$ (therefore the medians are close to 0.45) and they gradually converge to values which are estimated to provide the best update of the covariance matrix w.r.t. the ability to generate the current best individuals. It is seen that these values are problem and dimension-dependent. The values of $c_1$ are always much smaller than $c_{\mu}$ but are comparable to the default $c_1$. The values of $c_{\mu}$ and $c_c$ and $c_1$ are almost always larger than the default ones; this is not a surprise for $c_1$ and $c_{\mu}$, as their original default values are chosen in a rather conservative way to prevent degeneration of the covariance matrix. 

Several interesting observations can be made about the dynamics of the parameter values. 
The value of $c_{\mu}$ is high most of the times on the Rosenbrock functions, but it decreases toward values similar to those of the Sphere functions, when close to the optimum. 
This effect is observed on most problems; indeed, on most problems fast adaptation of the covariance matrix will improve the performance in the beginning, while the distribution shape should remain stable\del{it can be explained by a faster adaptation of the covariance matrix in the beginning and a Sphere-like search} when the covariance matrix is learned close to the optimum.

The overall performance of \Free-CMA-ES on the considered problems is comparable to that of CMA-ES, with a speed-up of a factor up to 1.5 on Sharp Ridge functions. The main result is the ability of \Free-CMA-ES to achieve the online adaptation of the hyper-parameters depending on the problem at hand, side-stepping the use of long calibrated default settings\footnote{$c_c = \frac{4}{n+4}$, $c_1 = \frac{2}{(n+1.3)^2 +\mu_w}$, $c_{\mu} = \frac{2 \, (\mu_w -2 + 1/{\mu_w})}{(n+2)^2+\mu_w}$.}.\niko{Why would one like to sidestep the use of something which has been carefully calibrated? } 
\ilya{Not sure whether the term "side-stepping" is best placed, but I think it might be nice to be show that without an initial approximation it is still possible to obtain good results. I agree that for the "real" algorithm, i.e., the one which should show good results right from the beginning and is not allowed to waste evaluations and/or on large-dimensional problems where the region of "good" parameters is small, the initial approximation will be used. Figure \ref{dampscs} shows the results for different initial step-size used for optimization of hyper-parameters damps and cs (note that all hyper-parameters are normalized to [0,1] when are optimized). The results suggests a role of the initial step-size. It seems that damps below 1 is better while the defaul one is never considered to be below 1 for any n and lambda. The main story here is that the results are not obtained for the original h function, but for a new one where one more step of CMA-ES is performed using points from the last generation.\niko{
I am not sure I do understand exactly what that means. } The same thing does not work for c1 and cmu since cmu then would be maximized to make C = Cmu. I am still quite confused about it. It seems that "optimal" damping gives almost 1.5 speed-up factor.}\niko{This is somewhat how the damping factor was tuned in the first place: conservative but without loosing out more than a factor of two. }


\subsection{Discussion}

\Free-CMA-ES offers a proof of concept for the online adaptation of three CMA-ES hyper-parameters in 
terms of feasibility and usefulness. Previous studies on parameter settings for CMA-ES mostly considered offline tuning (see, e.g., \cite{2010SmitEibenCMAEStuning,2013LiaoStutzleCEC}) and theoretical analysis dated back to the first papers on Evolution Strategies. The main limitation of these studies is that the suggested hyper-parameter values are usually specific to the (class of) analyzed problems. Furthermore, the suggested values are fixed, assuming that optimal parameter values remain constant along evolution. However, when optimizing a function whose landscape gradually changes when approaching the optimum, one may expect optimal hyper-parameter values to reflect this change as well. 

Studies on the online adaptation of hyper-parameters (apart from $\sigma$, $\vc{m}$ and $\vc{C}$)  
usually consider population size in noisy \cite{2013Beyer}, multi-modal \cite{2005AugerIPOP,2012LoshchilovNBIPOP} or expensive \cite{2006HoffmannAdaptLambdaCMAES} optimization. A more closely related approach was proposed in \cite{2012Schaulcomparing} where the learning rate for step-size adaptation is adapted in a stochastic way similarly to Rprop-updates \cite{2003Igelempirical}. 

\section{Conclusion and Perspectives}
\label{section:conclusion}

This paper proposes a principled approach for the self-adaptation of CMA-ES hyper-parameters, 
tackled as an \aux\ optimization problem: maximizing the likelihood of generating the best sampled solutions. The experimental validation of \Free-CMA-ES shows that the learning rates involved in 
the covariance matrix adaptation can be efficiently adapted on-line, with comparable or better 
results than CMA-ES. It is worth emphasizing that matching the performance of CMA-ES, the default setting of which represent a historical consensus between theoretical analysis and offline tuning, is nothing easy. 

The main novelty of the paper is to offer an intrinsic assessment of the algorithm internal 
state, based on retrospective reasoning (given the best current solutions, how could the generation of these solutions have been made easier) and on one assumption (the optimal hyper-parameter values at time $t$ are ''sufficiently good`` at time $t+1$). Further work will investigate how this intrinsic assessment can  support the self-adaptation of other continuous and discrete hyper-parameters used to deal with noisy, multi-modal and constrained optimization problems. 

 
\subsubsection*{Acknowledgments}

We acknowledge anonymous reviewers for their constructive comments. This work was supported by the grant ANR-2010-COSI-002 (SIMINOLE) of the French National Research Agency.

\bibliographystyle{abbrv}
\bibliography{ppsn2014}

\end{document}